\RequirePackage{fix-cm}
\documentclass[twocolumn]{svjour3}          
\smartqed  
\usepackage{graphicx}
\usepackage{epstopdf}
\usepackage{lineno,hyperref}
\usepackage{amssymb}
\usepackage{amsmath}
\usepackage{verbatim}
\usepackage[ruled,vlined]{algorithm2e}
\usepackage{microtype} 
\usepackage{bm}
\usepackage{listings}
\usepackage{comment}
\usepackage{float} 
\usepackage{xfrac}
\usepackage[all]{xy}
\usepackage{siunitx}
\usepackage{color}
\usepackage{nth}
\usepackage[ampersand]{easylist}

\journalname{KI - K\"unstliche Intelligenz}
\begin{document}

\title{Measuring the Quality of Explanations:\\
The System Causability Scale (SCS)
}
\subtitle{Comparing Human and Machine Explanations}

\titlerunning{System Causability Scale (SCS)}        

\author{Andreas Holzinger\\Andr\'e Carrington\\Heimo M\"uller
}

\authorrunning{Holzinger-Carrington-Mueller} 

\institute{Corresponding Author: Andreas Holzinger\\
ORCID: 0000-0002-6786-5194\\
Institute for Medical Informatics, Statistics \& Documentation, Medical University Graz, Austria, and xAI-Lab, Alberta Machine Intelligence Institute, Edmonton,~Canada \at
\email{andreas.holzinger@human-centered.ai}           \\
\\
Andr\'e Carrington\\
Clinical Epidemiology Program, Ottawa Hospital\\
Research Institute, Ottawa,~Canada \at
\email{acarrington@ohri.ca}\\
\\
Heimo M\"uller\\
Diagnostic and Research Institute of Pathology, \\
Medical University Graz, Austria \at
\email{heimo.mueller@mac.com}\\
}

\date{to appear}

\maketitle
\begin{abstract}
Recent success in Artificial Intelligence (AI) and Machine Learning (ML) allow problem solving automatically without any human intervention. Autonomous approaches can be very convenient. However, in certain domains, e.g., in the medical domain, it is necessary to enable a domain expert to understand, \textit{why} an algorithm came up with a certain result. Consequently, the field of Explainable AI (xAI) rapidly gained interest worldwide in various domains, particularly in medicine. Explainable AI studies transparency and traceability of opaque AI/ML and there are already a huge variety of methods. For example with layer-wise relevance propagation relevant parts of inputs to, and representations in, a neural network which caused a result, can be highlighted. This is a first important step to ensure that end users, e.g., medical professionals, assume responsibility for decision making with AI/ML and of interest to professionals and regulators. Interactive ML adds the component of human expertise to AI/ML processes by enabling them to re-enact and retrace AI/ML results, e.g. let them check it for plausibility. This requires new human-AI interfaces for explainable AI. In order to build effective and efficient interactive human-AI interfaces we have to deal with the question of \textit{how to evaluate the quality of explanations} given by an explainable AI system. In this paper we introduce our System Causability Scale (SCS) to measure the quality of explanations. It is based on our notion of Causability \cite{HolzingerEtAl:2019:Wiley-Paper} combined with concepts adapted from a widely-accepted usability scale.

\end{abstract}


\section{Introduction}

Artificial intelligence (AI) is an umbrella term for algorithms aiming at delivering task solving capabilities comparable to humans. A dominant sub-field is automatic (or autonomous) machine learning (aML) with the aim to develop software that can learn fully automatically from previous experience to make predictions based on new data. One currently very successful family of aML methods includes deep learning (DL), which is based on the concepts of neural networks, and the insight that the depth of such networks yields surprising capabilities.



Automatic approaches are present in daily practice of human society, supporting and enhancing our quality of life. A good example is the breakthrough achieved with DL~\cite{LeCunBengioHinton:2015:DeepLearningNature} on the task of phonetic classification for automatic speech recognition. Actually, speech recognition was the first commercially successful application of DL~\cite{HintonEtAl:2012:DeepSpeech}. Autonomous software is able today to conduct conversations with clients in call centers; Siri, Alexa and Cortana make suggestions to smartphone users. A further example is automatic game playing without human intervention~\cite{SilverHassabisEtAl:2017:GoWithoutHuman}. Mastering the game of Go has a long tradition and is a good benchmark for progress in automatic approaches, because Go is hard for computers~\cite{RichardsMikkulainen:1998:Go} because it is strategic, although games are a closed environment with clear rules and a large number of games can be simulated for big data.

Even in the medical domain, automatic approaches recently demonstrated impressive results: automatic image classification algorithms are on par with human experts or even outperforms them~\cite{EstevaThrun:2017:DermaNN}; automatic detection of pulmonary nodules in tomography scans detected the tumoral formations missed by the same human experts who provided the test data~\cite{SetioEtAl:2017:NoduleDetection}; neural networks outperformed a traditional segmentation methods~\cite{GhafoorianEtAl:2017:DeepWhite}, consequently, automatic deep learning approaches became quickly a method of choice for medical image analysis~\cite{LitjensEtAl:2017:DeepLearningSurveyMedImages}

Undoubtedly, automatic approaches are well motivated for theoretical, practical and commercial reasons. Unfortunately, such approaches have also several disadvantages. They are resource consuming, require much engineering effort, need large amounts of training data ("big data"), but most of all they are often considered as black-box approaches which do not foster trust and acceptance and most of all responsibility. International concerns are raised on ethical, legal and moral aspects of developments of AI in the last years, particularly in the medical domain \cite{WiensEtAl:2019:ResponsibleML}. One example of such international effort is the Declaration of Montreal~\footnote{https://www.montrealdeclaration-responsibleai.com}.

Lacking transparency means that such approaches do not expose explicitly the decision process \cite{CarringtonThesis}. This is due to the fact that such models have no explicit declarative knowledge representation, hence they have difficulty in generating the required explanatory structures – which considerably limits the achievement of their full potential~\cite{BolognaHayashi:2017:deep}.



Consequently, in the medical domain a human expert involved in the decision process can be beneficial yet mandatory~\cite{Holzinger:2016:human-in-the-loop}.
However, the problem is that many algorithms, e.g. deep learning, are inherently opaque, which causes difficulties both for the developers of the algorithms, as well as for the human-in-the-loop.

Understanding the reasons behind predictions, queries and recommendations \cite{CaleroEtAl:2016:HealthRecommender} is important for many reasons. Among the most important reasons is trust in the results which is improved by an explanatory interactive learning framework, where the algorithm is able to explain each step to the user and the user can interactively correct the explanation \cite{TesoKersting:2019:exiML}. The  advantage of this approach, called interactive machine learning (iML) \cite{Holzinger:2019:imL}, is to include the strengths of humans, in learning and explaining abstract concepts \cite{HolzingerEtAl:2019:KandinskyIQ}.




Current ML algorithms work asynchronously in connection with a human expert who is expected to help in data pre-processing (refer to \cite{HasslerEtAl:2019:frailty} for a recent example of the importance of data quality). Also the human is expected to help in data interpretation - either before or after the learning algorithm. The human expert is supposed to be aware of the problem's context and to correctly evaluate specific data sets. 

The iML-approaches can therefore be effective on problems with scarce and/or complex data sets, when aML methods become inefficient. Moreover, iML enables important mechanisms, including re-traceability, transparency and explainability, which are important characteristics for any future information system \cite{HolzingerKieseWeipplTjoa:2018:trends}. 

The efficiency and the effectiveness of explanations provided by ML and iML require further study \cite{doshi2017towards}.  One approach to the problem examines how people understand explanations from ML by qualitatively rating the effectiveness of three explanatory models
\cite{chander2018evaluating,lou2012intelligible}. Another approach measures a proxy for utility such as simplicity \cite{ribeiro2016should,CarringtonThesis} or response time in an application \cite{narayanan2018humans}. Our contribution is to directly measure the user's perception of an explanation's utility, including cause aspects, by adapting a well-accepted approach in usability \cite{Brooke:1996:SUS}.

\section{Causability and Explainability}
\label{sec:explainability}
\subsection{Definitions}

A statement $s$ (see Figure 1) is either be made by a human $s_h$ or a machine $s_m$.\\
\\
$s = f(r, k, c)$ is a function with the following parameters:

\begin{itemize}
\item [$r$] representations of an unknown (or unobserved) fact $u_e$ related to an entity, 
\item [$k$] pre-existing knowledge, which is for a machine embedded in an algorithm, or made up for human by explicit, implicit and  tacit knowledge, 
\item [$c$] context, for a machine the technical runtime environment, and for humans the physical environment the decision was made (pragmatic dimension).
\end{itemize}

An unknown (or unobserved) fact $u_e$ represents a ground truth $gt$ that we try to model with machines $m_m$ or as humans $m_h$.  Unobserved, hidden or latent variables are found in the literature for Bayesian models \cite{gelman2003fundamentals}, hidden Markov models \cite{fieguth2010statistical} and methods like probabilistic latent component analysis \cite{shashanka2008probabilistic}.


The overall goal is, that a statement is congruent with the \textbf{ground truth} and the explanation of a statement highlights applied parts of the model.

\begin{figure*}[tb]
	\centering
	\includegraphics[width=.99\textwidth]{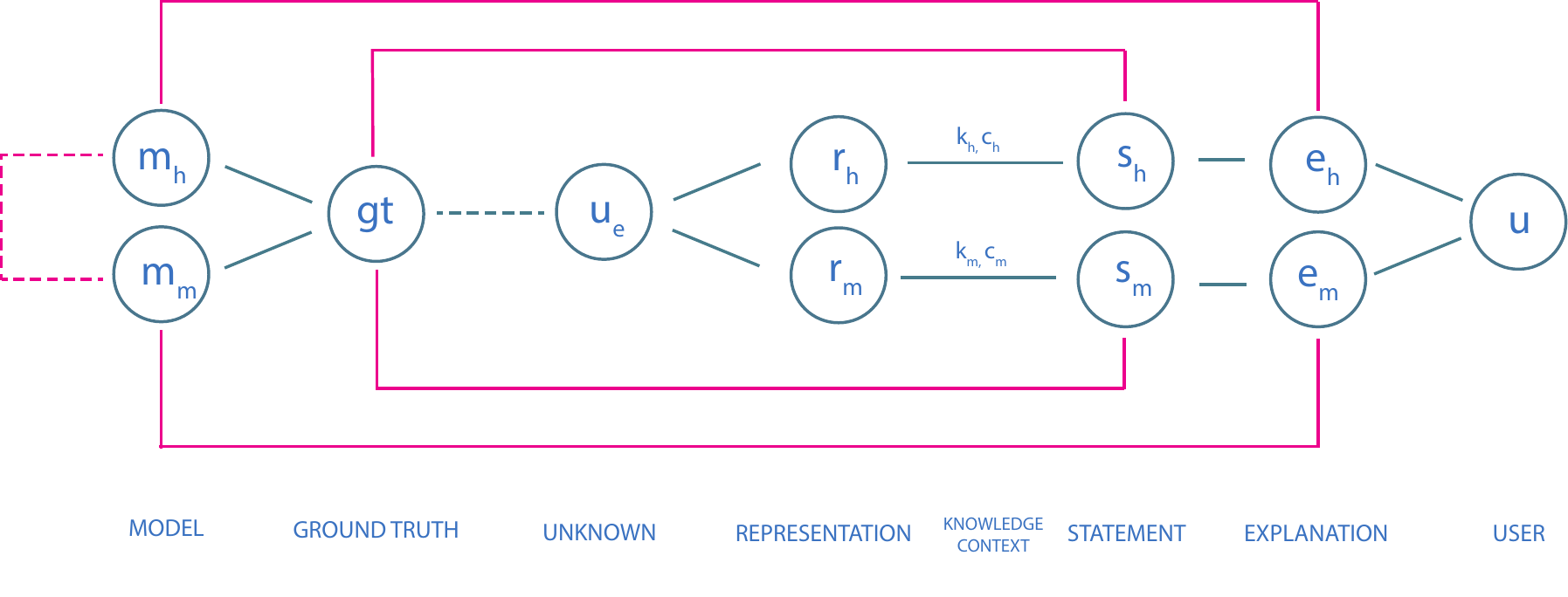}
	\caption{The Process of Explanation.  Explanations (e) by humans and machines (subscripts h and m) must be congruent with statements (s) and models (m) which in turn are based on the ground truth (gt).  Statements are a function of representations (r), knowledge (k) and context (c).}
	\label{fig:Figure-1}
\end{figure*}

\section{Process of Explanation and the importance of a Ground Truth}

In an ideal world the human and machine statement are identical, $s_h=s_m$, and congruent with the ground truth, which is defined for machines and humans within the same, $m_h=m_m$ (a connection between them, see Figure 1).

However, in the \textit{real world} we face two problems:\\ 
(i) ground truth is not always well defined, especially when making a medical diagnosis; and\\ 
(ii) although human (scientific) models are often based on understanding causal mechanisms, today's successful machine models or algorithms are typically based on correlation or related concepts of similarity and distance.\\  
\\
The latter approach in ML is probabilistic in nature and is viewed as an intermediate step which can only provide a basis for further establishing causal models. When discussing the explainability of a machine statement we therefore propose to distinguish between

\begin{itemize}
\item {Explainability, which in a technical sense highlights decision relevant parts of machine representations $r_m$ and machine models $m_m$---i.e., parts which contributed to model accuracy in training, or to a specific prediction. It does not refer to a human model $m_h$.}
\item {Causability \cite{HolzingerEtAl:2019:Wiley-Paper} as the extent to which an explanation of a statement to a user achieves a specified level of causal understanding with effectiveness, efficiency and satisfaction in a specified context of use.}
\end{itemize}

As causability is measured in terms of effectiveness, efficiency, satisfaction related to causal understanding and its transparency for a user, it refers to a human understandable model $m_h$. This is always possible for an explanation of a human  statement, as the explanation is per se defined related to $m_h$. 

To measure the causability of an explanation $e_m$ of a machine statement $s_m$ either $m_h$ has to be based on a causal model (which is not the case for most ML algorithms) or a mapping between $m_m$ and $m_h$ has to be defined.

\section{Background}


The System Usability Scale (SUS) has been in use for three decades and proved to be very efficient and necessary to rapidly determine the usability of a newly designed user interface. The SUS measures how usable a system's user-interface is, while our proposed System Causability Scale measures how useful explanations are and how usable the explanation interface is. 

The SUS was created by John Brooke already in 1986 when working at the Digital Equipment Corporation (DEC). 10 years later he published it as a book chapter \cite{Brooke:1996:SUS} which received (as of 01.10.2019) 7,949 citations on Google Scholar with an amazing trend upwards.

The success factor is \textit{simplicity:} SUS consists of a 10 item questionnaire, each item having five response options for the end-users. Consequently, it provides a “quick and dirty” tool for measuring the usability, which proofed to be very reliable \cite{BangorKortumMiller:2008:SUS-Eval}, and it is used for a wide variety of any products, not only user-interfaces \cite{Holzinger:2002:AAL}.

When a SUS is used, participants are asked to score the following 10 items with one of five responses that range from \textit{strongly agree} to \textit{strongly disagree}:

\begin{enumerate}
\item I think that I would like to use this system frequently.
\item I found the system unnecessarily complex.
\item I thought the system was easy to use.
\item I think that I would need the support of a technical person to be able to use this system.
\item I found the various functions in this system were well integrated.
\item I thought there was too much inconsistency in this system.
\item I would imagine that most people would learn to use this system very quickly.
\item I found the system very cumbersome to use.
\item I felt very confident using the system.
\item I needed to learn a lot of things before I could get going with this system
\end{enumerate}
   
Interpreting SUS scores can be difficult and one big disadvantage is that the scores (since they are on a scale from 0 to 100) are often wrongly interpreted as percentages. The best way to interpret results involves “normalizing” the scores to produce a percentile ranking. Consequently, the participant’s scores for each question are converted to a new number, added together and then multiplied by 2.5 to convert the original scores of 0-40 to 0-100. Though the scores are 0-100, these are not percentages and should be considered only in terms of their percentile ranking.

Based on a lot of research, a SUS score above 68 would be considered above average and anything below 68 is below average, however the best way to interpret the results involves “normalizing” the scores to produce a percentile ranking. 

A further disadvantage is that SUS has been assumed to be unidimensional. However, factor analysis of two independent SUS data sets reveals that the SUS actually has two factors – Usable (8 items) and Learnable (2 items – specifically, Items 4 and 10). 
These new scales have reasonable reliability (coefficient alpha of .91 and .70, respectively). They correlate highly with the overall SUS (r = .985 and .784, respectively) and correlate significantly with one another (r = .664), but at a low enough level to use as separate scales \cite{LewisSauro:2009:FactorSUS}. 

\section{The System Causability Scale}

In the following we propose our System Causability Scale (SCS) using the Likert scale similar to SUS. The Likert method \cite{Likert:1932:scale} is widely used as a standard psychometric scale to measure human responses (see about the limitations in the conclusions). 
The purpose of our SCS is to \textit{quickly} determine whether and to what extent an explainable user interface (human-AI interface), an explanation, or an explanation process itself is suitable for the intended purpose.


\begin{enumerate}
\item I found that the data included all relevant known causal factors with sufficient precision and granularity.
\item I understood the explanations within the context of my work.
\item I could change the level of detail on demand.
\item I did not need support to understand the explanations.
\item I found the explanations helped me to understand causality. 
\item I was able to use the explanations with my knowledge base.
\item I did not find inconsistencies between explanations.
\item I think that most people would learn to understand the explanations very quickly.
\item I did not need more references in the explanations: e.g., medical guidelines, regulations.
\item I received the explanations in a timely and efficient manner.
\end{enumerate}
As an illustration, SCS was applied by a medical doctor from the Ottawa Hospital (see the acknowledgement section) to the Framingham Risk Tool (FRT) \cite{genest2003recommendations}. FRT was selected as a classic example of a prediction model that is in use today.

FRT estimates the risk of coronary artery disease in 10 years for a patient without diabetes mellitus or clinically evident cardiovascular disease, and uses data from the Framingham Heart Study \cite{grundy1999assessment}. FRT includes the following input features: sex, age, total cholesterol
smoking, HDL (high density lipoprotein) cholesterol, systolic blood pressure and hypertension treatment. The ratings for the SCS score are reported in Table 1.

\begin{table}[h!]
\begin{center}
\caption{Using SCS with the Framingham Model. Ratings are: 1=strongly disagree, 2=disagree, 3=neutral, 4=agree, 5=strongly agree}
\begin{tabular}{|l|c|}
\hline 
Question & Rating 
\tabularnewline
\hline 
\hline 
01. Factors in data & 3
\tabularnewline
\hline 
02. Understood & 5 
\tabularnewline
\hline 
03. Change detail level & 5 \tabularnewline
\hline 
04. Need teacher/support & 5 \tabularnewline
\hline 
05. Understanding causality & 5 \tabularnewline
\hline 
06. Use with knowledge & 3 \tabularnewline
\hline 
07. No inconsistencies & 5 \tabularnewline
\hline 
08. Learn to understand & 3 \tabularnewline
\hline 
09. Needs references & 4 
\tabularnewline
\hline 
10. Efficient & 5 
\tabularnewline
\hline
\hline
\textbf{SCS} $ = \underset{i}\sum{Rating_i / 50} $ & \textbf{0.86}
\tabularnewline
\hline 
\end{tabular}
\end{center}
\end{table}

\section{Conclusions}

The purpose of the System Causability Scale is to provide a simple and rapid evaluation tool to measure the quality of an explanation interface (human-AI interface) or an explanation process itself. We were inspired by the System Usability Scale and the Framingham model which is often in use in daily routine.
The limitations of the SCS is that Likert scales fall within the ordinal level of measurement, meaning that the response categories have a rank order. However, the intervals between values cannot be presumed equal (it is illegitimate to infer that the intensity of feeling between strongly disagree and disagree is equivalent to the intensity of feeling between other consecutive categories on the Likert scale). The legitimacy of assuming an interval scale for Likert-type categories is an important issue, because the appropriate descriptive and inferential statistics differ for ordinal and interval variables and if the wrong statistical technique is used, the researcher increases the chance of coming to the wrong conclusion \cite{Jamieson:2004:Likert}. We are convinced that our Systems Causability Scale is useful for the international machine learning research community. Currently we are working on an evaluation study with the application in the medical domain.

\section*{Abbreviations}
AI ... Artificial Intelligence\\
aML ... automatic (or autonomous) Machine Learning\\
DL ... Deep Learning\\
FRT ... Framingham Risk Tool\\
iML ... interactive Machine Learning\\
ML ... Machine Learning\\
SCS ... System Causability Scale\\
SUS ... System Usability Scale

\section*{Acknowledgements}
The authors declare that there are no conflict of interests. This work does not raise any ethical issues. The authors are grateful for feedback and input from Dr. Douglas Manuel, MD, MSc, FRCPC from the Ottawa Health Research Institute and for comments from the international research community. Parts of this work have been funded by the Austrian Science Fund (FWF), Project:  P-32554 “A reference model of explainable Artificial Intelligence for the Medical Domain”.

\bibliographystyle{unsrt}
\bibliography{references}   


\end{document}